\pdfoutput=1

\documentclass[11pt]{article}

\usepackage{ACL2023}

\usepackage{times}
\usepackage{latexsym}
\usepackage{adjustbox}
\usepackage{amssymb}
\usepackage{amsmath}
\usepackage{multirow}

\newcommand\red[1]{\textcolor{red}{#1}}
\newcommand\blue[1]{\textcolor{blue}{#1}}

\usepackage[T1]{fontenc}

\usepackage[utf8]{inputenc}

\usepackage{microtype}

\usepackage{inconsolata}

\title{\textsf{Adapter-TST}: A Parameter Efficient Method for \\Multiple-Attribute Text Style Transfer}

\author{
    ~~Zhiqiang Hu$^{1}$
    ~~Roy Ka-Wei Lee$^{1}$
    ~~Nancy F. Chen$^{2}$\\ 
    $^{1}$Singapore University of Technology and Design, Singapore\\
    $^{2}$Institute of Infocomm Research (I2R), A*STAR, Singapore\\
}
\begin{document}
\maketitle
\begin{abstract}
Adapting a large language model for multiple-attribute text style transfer via fine-tuning can be challenging due to the significant amount of computational resources and labeled data required for the specific task. 
In this paper, we address this challenge by introducing \textsf{Adapter-TST}, a framework that freezes the pre-trained model's original parameters and enables the development of a multiple-attribute text style transfer model. Using BART as the backbone model, \textsf{Adapter-TST} utilizes different neural adapters to capture different attribute information, like a plug-in connected to BART. Our method allows control over multiple attributes, like sentiment, tense, voice, etc., and configures the adapters' architecture to generate multiple outputs respected to attributes or compositional editing on the same sentence. We evaluate the proposed model on both traditional sentiment transfer and multiple-attribute transfer tasks. The experiment results demonstrate that \textsf{Adapter-TST} outperforms all the state-of-the-art baselines with significantly lesser computational resources. We have also empirically shown that each adapter is able to capture specific stylistic attributes effectively and can be configured to perform compositional editing.
\end{abstract}

\section{Introduction}
\textbf{Motivation.} Text style transfer (TST) is a popular natural language generation task that aims to change the stylistic properties (e.g., sentiment, formality, tense, voice) of the text while preserving the style-independent content \cite{hu2022text}. Existing studies explore performing text style transfer on attributes like age, or gender \cite{lampleSSDRB19}, sentiment \cite{li2018delete,Luo19DualRL,fu2018style}, formality \cite{rao2018dear}, politeness \cite{madaan-etal-2020-politeness,hu2022current}, and author writing style \cite{syed2020adapting}. Nevertheless, most of the existing TST studies are confined to single-attribute TST tasks.

\begin{figure}[t] 
	\centering
	\includegraphics[scale = 0.14]{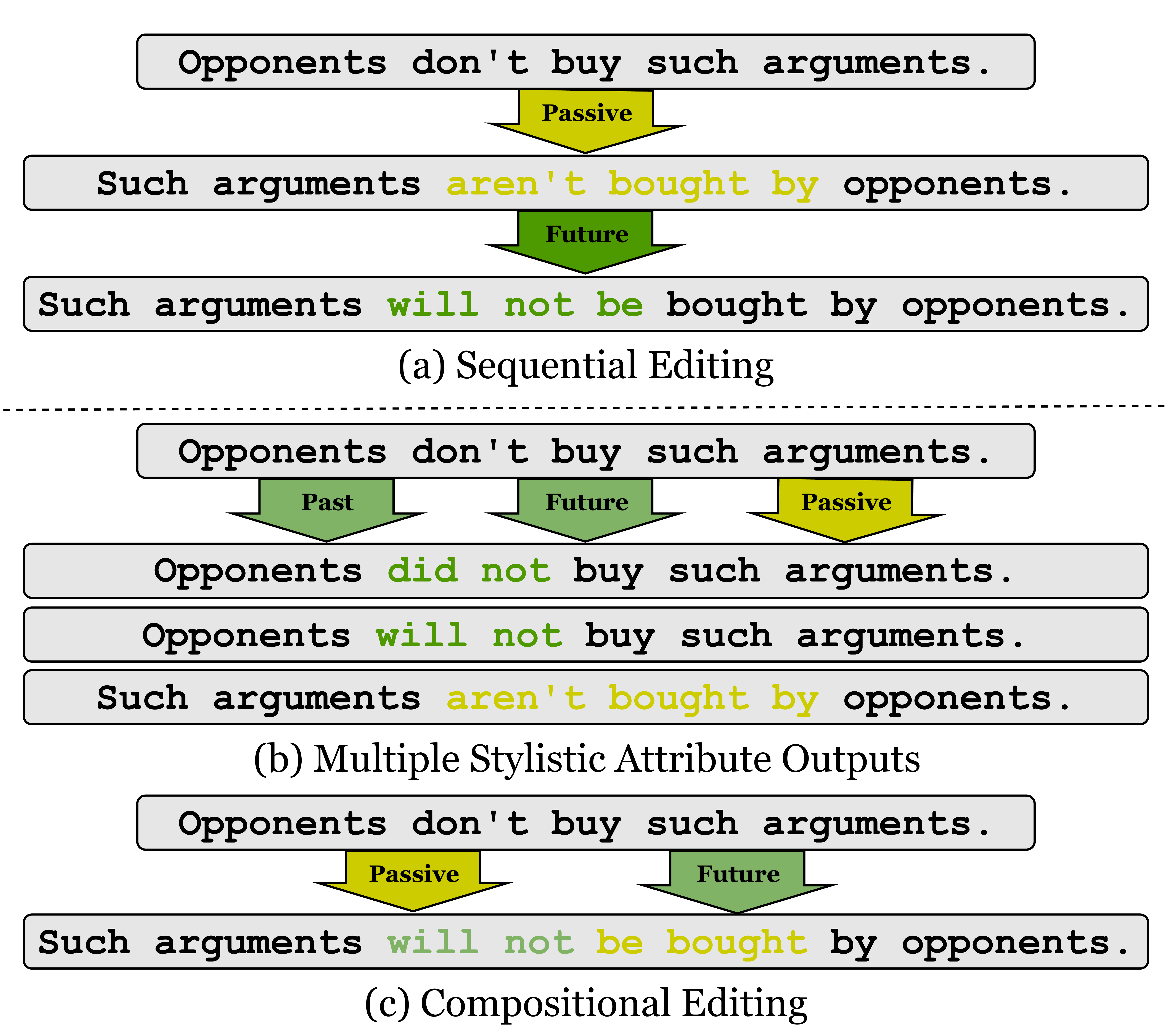} 
	\caption{Examples of different settings of multiple-attribute TST. (a) Existing single-attribute TST models perform sequential editing by transferring the text style sequentially to achieve compositional editing. Multiple-attribute TST models can (b) generate multiple outputs simultaneously in the corresponding target style, or (c) perform compositional editing by transferring different target styles. The proposed \textsf{Adapter-TST} enables a single PLM to achieve both settings (b) and (c) by configuring the adapters' connection method.}
	\label{fig:example}
\end{figure}

Few works have explored multiple-attribute TST tasks, where TST models are designed to control and transfer text in multiple target stylistic attributes. \citet{lampleSSDRB19} attempts style transfer with multiple attributes by conditioning on the average embedding of each target attribute and using a combination of denoising autoencoder (DAE) and back-translation techniques. \citet{goyal-etal-2021-multi} propose an approach to initialize an encoder-decoder setup with a transformer-based language model that is pre-trained on a generic corpus and enhances its capability of re-writing to multiple target style dimensions by utilizing multiple style-aware language models as discriminators. 

A possible approach to perform single and multiple attribute TST tasks is to leverage large pre-trained language models (PLMs). The PLMs have been pre-trained on large corpora, which allows them to capture natural language's syntactic and semantic information. This characteristic of PLMs makes them well-suited for TST tasks, where the model needs to understand the content and style of the input text. \citet{syed2020adapting} fine-tune a denoising autoencoder (DAE) for the stylized re-writing task by initializing the encoder and decoder with a pre-trained language model trained on Masked Language Modeling (MLM) objectives \cite{devlin-etal-2019-bert}. \citet{wang2019harnessing} fine-tune GPT-2 model \cite{radford2019language} using the text formality transfer rules harnessed from analyzing the GYAFC parallel dataset \cite{rao2018dear}. The fine-tuned GPT-2 model was subsequently used to transfer the formality of text (e.g., informal to formal text). However, fine-tuning PLMs for multiple-attribute TST remains challenging as a significant amount of computational resources and style-labeled data are required to perform TST for each stylistic attribute.

\textbf{Research Objectives.} To address these research gaps, we propose \textsf{Adapter-TST}, a parameter-efficient framework that utilizes BART \cite{lewis-etal-2020-bart} as the backbone model and trains neural adapters to capture multiple stylistic attributes for multiple-attribute TST. During the training of \textsf{Adapter-TST}, we freeze the original parameters of the pre-trained BART model and only update the parameters of adapters to relax the dependence on computational resources and supervised data. The proposed \textsf{Adapter-TST} model is flexible to handle different settings of multiple-attribute TST by configuring the connection method among adapters. Figure \ref{fig:example} illustrates the different settings of multiple-attribute TST tasks. Paralleling the adapters in \textsf{Adapter-TST} can generate multiple outputs in the corresponding target style simultaneously (setting b) and stacking the adapters for compositional editing in terms of different target styles at the same time (setting c). We conduct experiments on the traditional sentiment transfer task and multiple-attribute TST tasks, including multiple stylistic attribute outputs and compositional editing. Results of automatic and human evaluations show that \textsf{Adapter-TST} can outperform the state-of-the-art baselines to transfer and generate high-quality text with lower computational resources. 

\textbf{Contributions.} We summarize our contributions as follows: (i) We introduce an \textsf{Adapter-TST}, which is a parameter-efficient framework that can perform multiple-attribute TST tasks with significantly lower computational resources. (ii) Included in the \textsf{Adapter-TST} are two TST configurations, \textit{parallel} and \textit{stacking}, which support  multiple-output TST and compositional editing, respectively. (iii) We conducted extensive experiments on real-world datasets. The automatic and human evaluation results show that \textsf{Adapter-TST} can outperform the state-of-the-art baselines to transfer and generate high-quality text.


\section{Related Work}
\subsection{Text Style Transfer}
TST is an emerging research topic that has garnered attention from computational linguists and computer science researchers. The recent comprehensive survey \cite{hu2022text,jin-etal-2022-deep} summarizes the existing TST approaches. 

While the majority of existing studies have focused on performing TST on single attributes such as sentiment~\cite{li2018delete,Luo19DualRL,fu2018style} or formality~\cite{rao2018dear}, recent studies have also explored multiple-attribute TST tasks, where TST models are designed to control and transfer text in multiple target stylistic attributes. \citet{lampleSSDRB19} attempts style transfer with multiple attributes by conditioning on the average embedding of each target attribute and using a combination of denoising autoencoder (DAE) and back-translation techniques. \citet{goyal-etal-2021-multi} propose an approach to initialize an encoder-decoder setup with a transformer-based language model that is pre-trained on a generic corpus and enhances its capability of re-writing to multiple target style dimensions by utilizing multiple style-aware language models as discriminators. In this study, we contribute to this limited multiple-attribute TST literature by proposing an alternative approach to generate multiple stylistic outputs and perform compositional editing efficiently. 

Due to the lack of parallel training data, most existing TST methods are designed to train with non-parallel style-labeled sentences as input. A popular line of TST approaches aims to disentangle the text's content and style in the latent space to perform TST~\cite{shen2017style,zhao2018adversarially,fu2018style,chen2018adversarial,logeswaran2018content,yin2019utilizing,lai2019multiple,john2019disentangled}. Another common approach is to leverage PLMs. For instance, \citet{syed2020adapting} fine-tune a denoising autoencoder (DAE) for the stylized re-writing task by initializing the encoder and decoder with a pre-trained language model trained on Masked Language Modeling (MLM) objectives \cite{devlin-etal-2019-bert}. \citet{wang2019harnessing} fine-tune GPT-2 model \cite{radford2019language} using the text formality transfer rules harnessed from analyzing the GYAFC parallel dataset \cite{rao2018dear}. The fine-tuned GPT-2 model was subsequently used to transfer the formality of text (e.g., informal to formal text). However, fine-tuning PLMs for multiple-attribute TST remains challenging as a significant amount of computational resources is required to perform the task; multiple PLMs need to be fine-tuned for the different attributes to perform multiple-attribute TST. In this study, we overcome this limitation by proposing \textsf{Adapter-TST}, which is a parameter-efficient framework that leverages on PLMs but requires significantly lesser computational resources to perform multiple-attribute TST.

\subsection{Adapter-based Models}
PLMs, pre-trained on large-scale text corpus with unsupervised objectives, have established state-of-the-art performances on various NLP downstream tasks. Many studies fine-tune PLMs with language modeling and downstream task objectives to obtain better performance \cite{zhang-etal-2019-ernie,lauscher2019informing,he2019integrating,xiong2019pretrained}.To leverage the powerful PLMs more efficiently, \citet{houlsby2019parameter} add adapter layers, small neural networks, into each transformer layer to obtain near state-of-the-art performance on the GLUE benchmark while updating only the parameters of adapters. Inspired by this work, more adapter-based models \cite{wang-etal-2021-k,liu-etal-2021-lexicon,zhong-etal-2021-useradapter} are proposed to inject task-specific knowledge into PLMs with adapters. Inspired by the adapter architecture, we propose \textsf{Adapter-TST}, which trains different neural adapters to capture different stylistic attributes to perform the multiple-attribute TST. The proposed adapter framework has two configurations that support multiple stylistic attribute outputs and compositional editing.

\section{Methodology}
This section proposes \textsf{Adapter-TST}, which adds neural adapters into each transformer layer to capture different attribute information for multiple-attribute TST. We first introduce the adapter structure used in \textsf{Adapter-TST} and its parameter efficiency. Subsequently, we explain how the adapters are configured for different multiple-attribute TST settings, namely, \textit{multiple stylistic attribute outputs} and \textit{compositional editing}. Finally, we describe the training objectives of \textsf{Adapter-TST}.

\subsection{Adapter Structure}

\begin{figure}[t] 
	\centering
	\includegraphics[scale = 0.15]{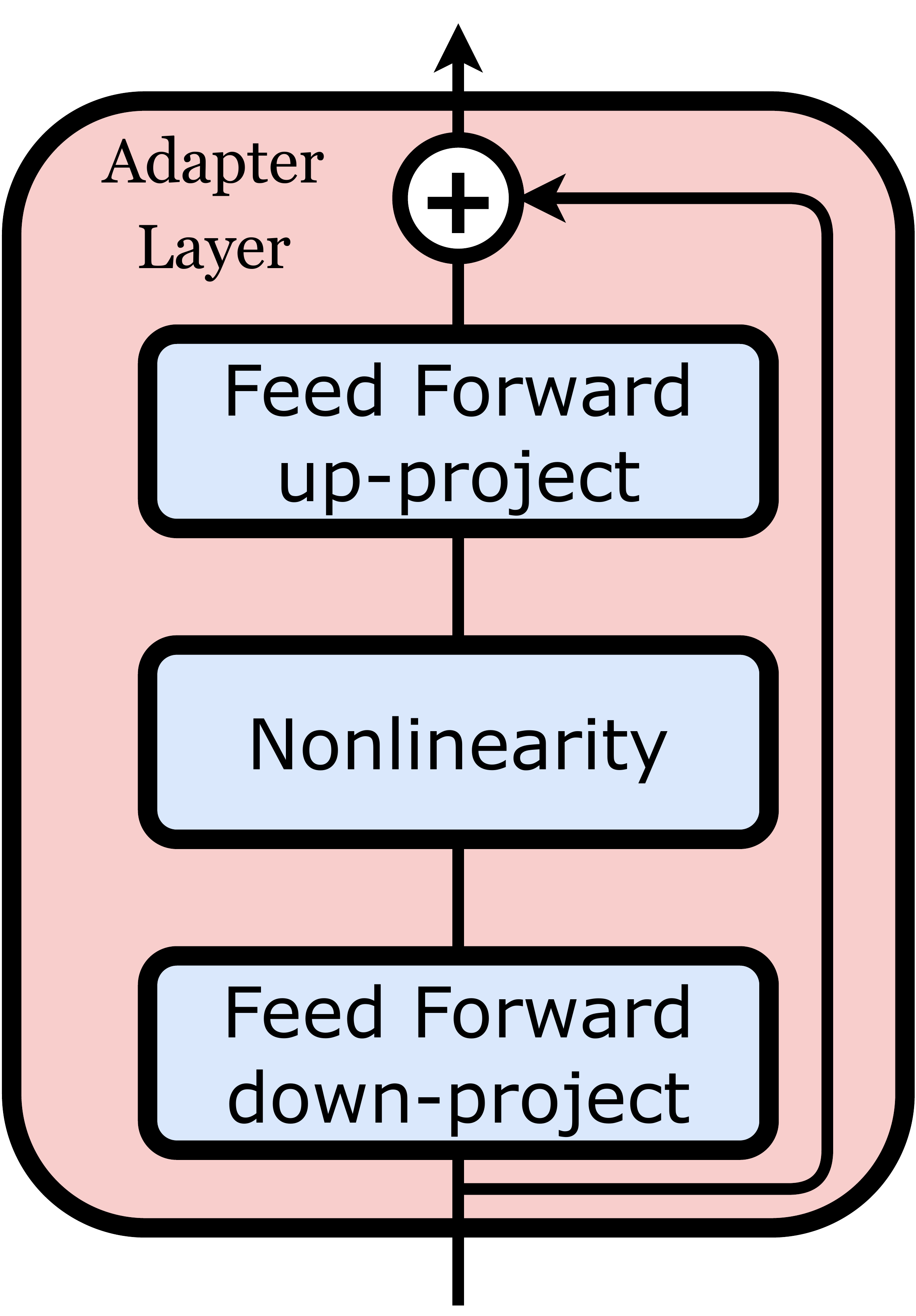} 
	\caption{Structure of the adapter layer. The adapter layer consists of a bottleneck with up and down projection layers, and a skip connection between two projection layer.}
	\label{fig:adapter_layer}
\end{figure}

\begin{figure*}[t] 
	\centering
	\includegraphics[scale = 0.09]{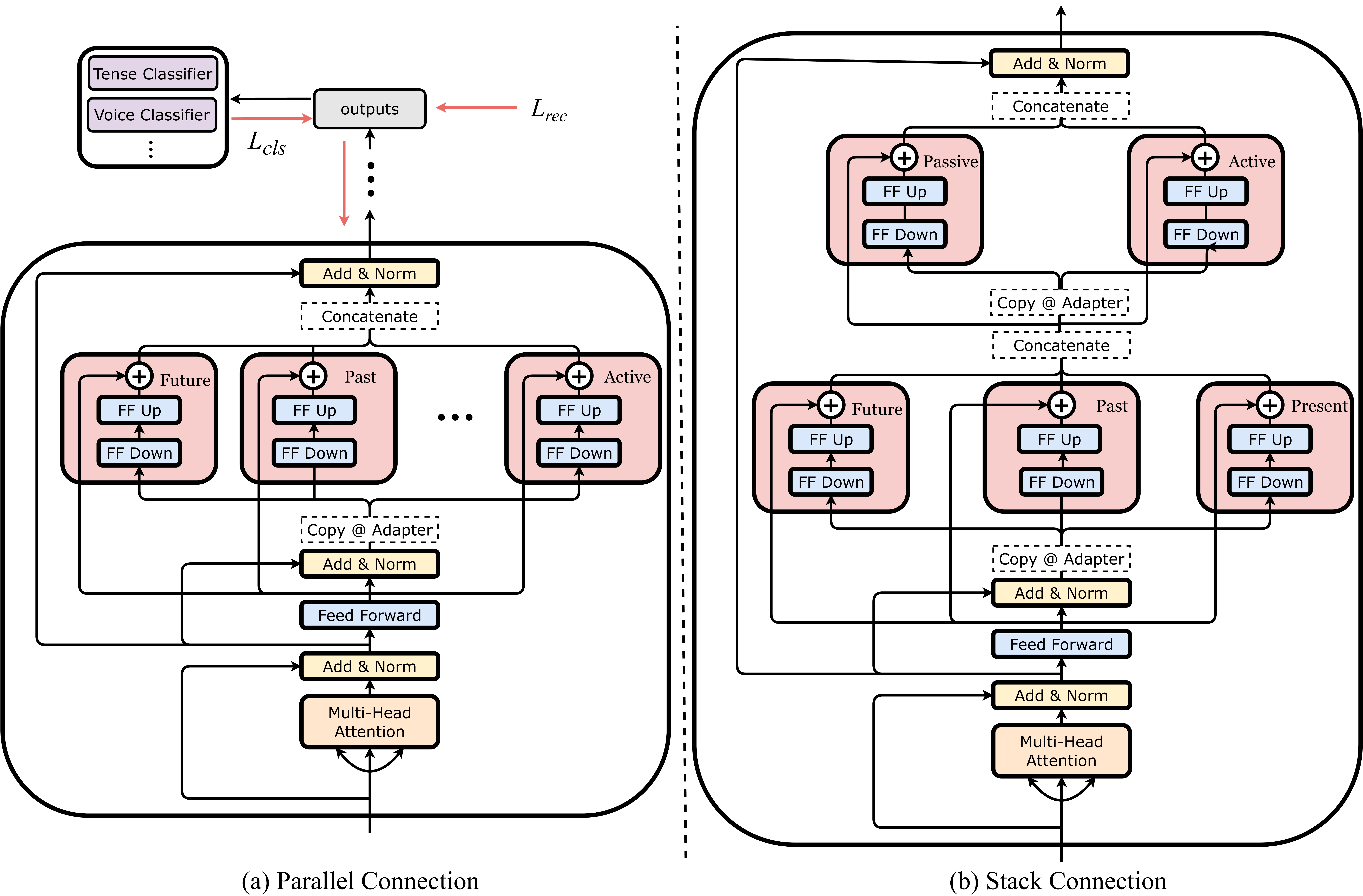} 
	\caption{\textsf{Adapter-TST} Configurations - Left: Paralleling the adapters enables a single PLM to model different attributes simultaneously and generate multiple outputs in the corresponding target style. Right: Stacking the adapters for compositional editing in terms of different target styles at the same time. \textit{Stack} connection is used for inference to verify the relevant attribute information captured by adapters.}
	\label{fig:model}
\end{figure*}

We present an adapter structure in Figure \ref{fig:adapter_layer}. The adapter consists of a bottleneck that contains few parameters relative to the attention and feedforward layers in the original model. A skip connection is applied across two projection layers. In our proposed \textsf{Adapter-TST}, these adapters will be trained to capture different stylistic attributes. In contrast to \citet{houlsby2019parameter}, adding the adapter module twice to each transformer layer, we propose simplifying the approach by just adding the adapter layer into each transformer once, making our \textsf{Adapter-TST}'s architecture more parameter efficient. 

We use BART-large (24-layer, 1024-hidden, 16-heads, 406M parameters) as the backbone model in \textsf{Adapter-TST}. As for each adapter layer, we denote the hidden dimensions of the down-projection and up-projection layers as $H_d = 64$ and $H_u = 1024$. The bottleneck adapter layers are plugged into each layer of BART-large, and different adapter layers do not share parameters. Thus the total number of parameters for each attribute-specific adapter is about 3.17M, which is only 0.78\% of the original BART-large model, making the training process parameter efficient. Note that the original parameters of BART-large are frozen during multiple-attribute TST training, and only the parameters of adapters are trainable and initialized randomly. 

\subsection{Adapter-TST Configurations}
\textsf{Adapter-TST} has two configurations, \textit{parallel} and \textit{stack}, which support two multiple-attribute TST task settings: \textit{multiple stylistic attribute outputs} and \textit{compositional editing}, respectively. To better understand the two Configurations of \textsf{Adapter-TST}, we take the multiple-attribute TST task with \textit{tense} and \textit{voice} attributes as an example. \textit{Tense} has three attribute values (\textit{Future}, \textit{Past}, \textit{Present}), while \textit{voice} has two attribute values (\textit{Passive}, \textit{Active}). Thus, we add five attribute-specific adapters \textit{Adapter(Future, Past, Present, Passive, Active)} to the BART-large model for all the possible attribute values, respectively. Each adapter is employed to learn to generate sentences with corresponding attribute values while preserving the semantic content of the inputs. 

\textbf{Parallel Connection.} We define the multiple stylistic attribute outputs as follows: given a sentence $x = \{x_1, ..., x_n\}$ with $n$ tokens and $y_{tense}, y_{voice}$ labels, the \textsf{Adapter-TST} model is required to generate multiple outputs with all possible other attribute values at the same time. For instance, as shown in Figure \ref{fig:example}(b), given an input sentence with present tense and active voice, the multiple-attribute TST models need to generate three sentences in the past tense, future tense, and passive voice simultaneously. The multiple stylistic attribute output setting requires TST models to capture all the stylistic attributes and have the capability of performing style transfer among the attribute values. \textsf{Adapter-TST} performs the multiple stylistic attribute output by utilizing the \textit{Parallel} connection configuration shown in Figure \ref{fig:model}(a). Specifically, we plug the paralleled adapters \textit{Parallel(Future, Past, Present, Passive, Active)} into each transformer layer of the BART-large model. During training, each training sample passes all the attribute-specific adapters, but adapters will take different actions according to the attribute values of input sentences. The adapter learns to reconstruct the input sentence for training samples with the same attribute value as an adapter. Conversely, when training samples with different attribute values, the adapter learns to transfer the attribute of the input sentence while preserving the original content. The outputs of all the adapters are concatenated together to the next layer. The replication is only performed once in the first transformer layer. In the latter transformer layers, we distribute the hidden states to corresponding adapters to make sure that the input of an adapter in the current layer is the output of the adapter with the same attribute value in the preceding layer.

\textbf{Stack Connection.} Compositional editing requires TST models to change multiple attributes simultaneously while preserving the original content. For instance, as shown in Figure \ref{fig:example}(c), given an input sentence with present tense and active voice, the multiple-attribute TST models need to generate one sentence both in future tense and passive voice. 
\textsf{Adapter-TST} performs compositional editing by using the \textit{Stack} connection method shown in Figure \ref{fig:model}(b), where adapters belonging to the same attribute are parallelized because a sentence should only contain one attribute value for a specific attribute. Specifically, we have \textit{Parallel(Future, Past, Present)} and \textit{Parallel(Passive, Active)} for tense and voice attributes. The two sets of paralleled adapters are stacked as \textit{Stack(Parallel(Future, Past, Present), Parallel(Passive, Active))} to learn to transfer multiple attributes. Similar to the \textit{Parallel} connection method, the hidden states are replicated according to the number of adapters in the \textit{Parallel} connection module. It's worth noting that, to demonstrate the attribute-specific adapters captured the attribute information, we only use the \textit{Stack} connection method in inference time. During inference, we reload the parameters of adapters trained in multiple stylistic attribute outputs tasks and change the connection among the adapters to \textit{Stack}. 

\subsection{Training Objectives.} 
TST tasks have two objectives, transferring the style of inputs while preserving the original semantic content. Thus, we train \textsf{Adapter-TST} with classification loss $L_{cls}$ for style transfer and reconstruction $L_{rec}$ for content preservation. During training, the original parameters of BART-large are frozen, and only the parameters of adapters are trainable. 

\textbf{Classification Loss $L_{cls}$}: The classification loss ensures that the transferred sentence conforms to the target attribute value. To this end, we first pre-train a TextCNN-based \cite{kim2014convolutional} binary attribute classifier $D$ for each attribute, then apply the pre-trained attribute classifiers to guide the updates of adapters' parameters such that the output sentence is predicted to be in the target style: 

\begin{equation}
    L_{cls} = - \mathbb{E}_{(x,y) \sim D}[logP(y_t|x')]
\end{equation}

where $x'$ is sampled from the distribution of model outputs at each decoding time step, $y_t$ is the target attribute value. We employ the policy gradient algorithm \cite{sutton1999policy} to deal with the discrete training issue with the attribute classifiers. 

\textbf{Reconstruction Loss $L_{rec}$}: The reconstruction loss attempts to preserve the original content information in the transferred sentences. Specifically, the loss function constricts the adapters to capture informative features to reconstruct the original sentence using the learned representations. Formally, we define $L_{rec}$ as follows:

\begin{equation}
    L_{rec} = -logP(x|z_i,y_i)
\end{equation}

where $y_i$ is the $i$-th attribute value of the input sentence, $z_i$ denotes the hidden representation extracted by the corresponding adapter. The input sentences are only reconstructed by the corresponding adapter and transferred by other adapters.

\textbf{Putting them together,} the final joint training loss $L$ is as follows:

\begin{equation}
    L = (1 - \lambda) L_{rec} + \lambda L_{cls}
\end{equation}

Where $\lambda$ is a balancing hyper-parameter to ensure that the transferred sentence has the target style while preserving the original content.

\section{Experiments}
\subsection{Experiment Setting}
\textbf{Datasets.} We evaluate the proposed \textsf{Adapter-TST} model on sentiment transfer  and multiple-attribute TST tasks using the Yelp\footnote{https://github.com/luofuli/DualRL} and StylePTB \cite{lyu-etal-2021-styleptb} datasets, respectively. We adopt the train, development, and test split for the Yelp dataset as \cite{Luo19DualRL}. \citet{lyu-etal-2021-styleptb} introduce StylePTB \footnote{https://github.com/lvyiwei1/StylePTB}, a large-scale benchmark with compositions of multiple-attribute TST tasks which allow the modeling of fine-grained stylistic changes. In our experiments, we choose four subsets for multiple-attribute TST: Tense-Voice, Tense-PP-Front$\leftrightarrow$Back, Tense--PP-Removal, and Tense-ADJADV-Removal. Specifically, the four subsets include five attributes, tense with three attribute values (\textit{Future}, \textit{Past}, \textit{Present}), voice with two attribute values (\textit{Passive}, \textit{Active}), proposition position with two attribute values (\textit{Front}, \textit{Back}),  proposition removal with two attribute values (\textit{Adding}, \textit{Removing}), and adjectives\&adverbs removal with two attribute values (\textit{Adding}, \textit{Removing}). Table~\ref{tbl:dataset_statistics}  shows the training, validation, and test splits of the Yelp and StylePTB datasets used in our experiments.

\begin{table}
\small
\centering
\begin{tabular}{c|c|c|c}
\hline
\textbf{Dataset} &  \textbf{Train} & \textbf{Dev} & \textbf{Test} \\
\hline\hline
Yelp  & 443K & 1,000 & 1,000 \\
Tense-voice & 28K & 1,538 & 1,564 \\
Tense-PP-Front$\leftrightarrow$Back & 5K & 270 & 284 \\
Tense-PP-Removal & 32K & 1,796 & 1,834 \\
Tense-ADJADV-Removal & 33K & 1,838 & 1,819 \\
\hline
\end{tabular}
\caption{Dataset statistics for Yelp and StylePTB.}
\label{tbl:dataset_statistics}
\end{table}

\textbf{Baselines.} For sentiment transfer, we benchmark \textsf{Adapter-TST} against nine state-of-the-art TST models: \textit{BackTrans} \cite{prabhumoye2018style}, \textit{CrossAlign} \cite{shen2017style}, \textit{DualRL} \cite{Luo19DualRL}, \textit{Unpaired} \cite{li2019domain}, \textit{UnsuperMT} \cite{zhang2018style}, \textit{Style Transformer} \cite{dai2019style}, \textit{DeleteOnly}, \textit{Template}, and \textit{Del\&Retri} \cite{li2018delete}. For multiple stylistic attribute outputs task, \textit{Style Transformer} \cite{dai2019style}, a transformer-based model for single-attribute TST, is selected as a baseline. We train multiple \textit{Style Transformer} models for each attribute and perform style transfer separately. For compositional editing, we use the trained \textit{Style Transformer} models to perform sequential editing, which transfers one attribute after another to compare results with our model. We term this baseline as \textit{Sequential Style Transformer} setup. 

\textbf{Training.} The experiments were performed on an Ubuntu 20.04.3 LTS system with 24 cores, 128 GB RAM, and Nvidia RTX 3090. The model implementation is based on AdapterHub \cite{pfeiffer2020AdapterHub} and Huggingface Transformers \cite{wolf-etal-2020-transformers}. For the balancing hyper-parameter $\lambda$, we choose the best-performed one from (0.9, 1) as the BART-large model can copy the input without training with TST objectives. 

\subsection{Automatic Evaluation}
We evaluate the proposed model and baselines on three criteria commonly used in TST studies: \textit{transfer strength}, \textit{content preservation}, and \textit{fluency}. An attribute classifier is first pre-trained to predict the attribute label of the input sentence. The classifier is subsequently used to approximate the style transfer accuracy (ACC) of the sentences’ transferred attributes by considering the target attribute value as the ground truth. To quantitatively measure the amount of original content preserved after style transfer operations, we employ BERTscore \cite{zhang2019bertscore} between style-transferred and original sentences. For fluency, We use GPT-2 \cite{radford2019language} to measure the perplexity (PPL) of transferred sentences. The sentences with smaller PPL scores are considered more fluent. Finally, we compute the geometric mean of ACC, BERTscore, and 1/PPL. We take the inverse of the calculated perplexity score because a smaller PPL score corresponds to better fluency. When there is more than one accuracy in the multiple-attribute TST tasks, we use the average accuracy to compute G-score. 


\begin{table}[t]
\small
\centering
\begin{tabular}{ccccc}
\hline
\textbf{Model} & \textbf{ACC}  & \textbf{BS} & \textbf{PPL} & \textbf{G} \\
\hline
BackTrans & 94.5 & 0.88 & 11.3 & 1.95 \\
CrossAlign & 74.3 & 0.89 & 35.3 & 1.23 \\
DeleteOnly & 87.6 & 0.91 & 36.4 & 1.30 \\
Del\&Retri & 90.2 & 0.91 & 34.0 & 1.34 \\
DualRL & 88.9 & \textbf{0.95} & 27.1 & 1.46 \\
Template & 83.7 & 0.92 & 47.2 & 1.18 \\
Unpaired & 50.6 & 0.91 & 53.1 & 0.95 \\
UnsuperMT & \textbf{96.2} & 0.93 & 33.5 & 1.39 \\
Style Transformer & 85.8 & \textbf{0.95} & 10.1 & 2.00 \\
\hline
\textsf{Adapter-TST} & 90.1 & 0.91 & \textbf{8.2} & \textbf{2.15} \\
\hline
\end{tabular}
\caption{Performance of models on Yelp dataset (Sentiment Transfer Task). The best performances are \textbf{bold}. }
\label{tbl:yelp_results}
\end{table}

\begin{table*}[t]
\small
\centering
\begin{tabular}{cc|ccccc|ccccc}
\hline
& & \multicolumn{5}{c}{Tense-Voice} & \multicolumn{5}{|c}{Tense-ADJADV-Removal} \\
\textbf{Model} & \textbf{Attri} & \textbf{Tense}  & \textbf{Voice}  & \textbf{BS} & \textbf{PPL} & \textbf{G} & \textbf{Tense}  & \textbf{Removal}  & \textbf{BS} & \textbf{PPL} & \textbf{G} \\
\hline
Style Transformer & single & 91.1 & - & 0.91 & 15.3 & 1.76 & 92.6 & - & 0.92 & 27.0 & 1.47 \\
Style Transformer & single & - & \textbf{87.2} & 0.85 & 11 & 1.89 & - & \textbf{83.7} & 0.93 & 21.7 & 1.53 \\
\textsf{Adapter-TST} (ours) & multi & \textbf{96.9} & 81.9 & \textbf{0.96} & \textbf{4.7} & \textbf{2.63} & \textbf{96.2} & 76.5 & \textbf{0.95} & \textbf{11.8} & \textbf{1.91} \\
\hline
& & \multicolumn{5}{c}{Tense-PP-Front$\leftrightarrow$Back} & \multicolumn{5}{|c}{Tense-PP-Removal} \\
\textbf{Model} & \textbf{Attri} & \textbf{Tense}  & \textbf{F$\leftrightarrow$Back}  & \textbf{BS} & \textbf{PPL} & \textbf{G} & \textbf{Tense}  & \textbf{Removal}  & \textbf{BS} & \textbf{PPL} & \textbf{G} \\
\hline
Style Transformer & single & \textbf{95.}7 & - & 0.83 & 6.8 & 2.27 & 94.9 & - & 0.91 & 27 & 1.47 \\
Style Transformer & single & - & \textbf{57.2} & 0.83 & 10.4 & 1.66 & - & \textbf{87.2} & 0.91 & 26.1 & 1.45 \\
\textsf{Adapter-TST} (ours) & multi & 88.2 & 48.9 & \textbf{0.96} & \textbf{4} & \textbf{2.54} & \textbf{96} & 74.5 & \textbf{0.96} & \textbf{12.5} & \textbf{1.87} \\
\hline
\end{tabular}
\caption{Automatic evaluation results of models on multiple stylistic attribute outputs task. The best performances are \textbf{bold}.}
\label{tbl:multi_output_results}
\end{table*}

\begin{table*}[t]
\small
\centering
\begin{tabular}{c|ccccc|ccccc}
\hline
& \multicolumn{5}{c}{Tense-Voice} & \multicolumn{5}{|c}{Tense-ADJADV-Removal} \\
\textbf{Model}  & \textbf{Tense}  & \textbf{Voice}  & \textbf{BS} & \textbf{PPL} & \textbf{G} & \textbf{Tense}  & \textbf{Removal}  & \textbf{BS} & \textbf{PPL} & \textbf{G} \\
\hline
Sequential Style Transformer & 80.2 & \textbf{88.1} & 0.85 & 22.2 & 1.48 & 88.6 & 90.0 & \textbf{0.89} & 42.2 & 1.23 \\
\textsf{Adapter-TST} (ours) & \textbf{88.2} & 85.4 & \textbf{0.90} & \textbf{8.0} & \textbf{2.14} & \textbf{88.9} & \textbf{92.7} & 0.86 & \textbf{22} & \textbf{1.53} \\
\hline
& \multicolumn{5}{c}{Tense-PP-Front$\leftrightarrow$Back} & \multicolumn{5}{|c}{Tense-PP-Removal} \\
\textbf{Model} & \textbf{Tense}  & \textbf{F$\leftrightarrow$Back}  & \textbf{BS} & \textbf{PPL} & \textbf{G} & \textbf{Tense}  & \textbf{Removal}  & \textbf{BS} & \textbf{PPL} & \textbf{G} \\
\hline
Sequential Style Transformer & 76.1 & \textbf{65.7} & 0.82 & 8.1 & 1.93 & \textbf{91.2} & 85.7 & \textbf{0.88} & 51.4 & 1.15\\
\textsf{Adapter-TST} (ours) &  \textbf{88.2} & 50.0 & \textbf{0.92} & \textbf{4.9} & \textbf{2.35} & 90.1 & \textbf{88.2} & 0.86 & \textbf{20.9} & \textbf{1.54} \\
\hline
\end{tabular}
\caption{Automatic evaluation results of models on compositional editing task. The best performances are \textbf{bold}.}
\label{tbl:compsotional_results}
\end{table*}

\subsection{Automatic Evaluation Results}
Table \ref{tbl:yelp_results} shows the performance of the \textsf{Adapter-TST} model and the baselines on the sentiment transfer task. \textsf{Adapter-TST} has achieved the best G-score, outperforming the state-of-the-art baselines. We observe that \textsf{Adapter-TST} achieves comparable performance on transfer strength and content preservation with 90.1\% transfer accuracy and 0.91 BERTscore by only updating the parameters of adapters. With the impressive generative ability of the pre-trained BART model, the \textsf{Adapter-TST} model can generate high-quality text in terms of fluency and completeness. The experiment results demonstrate \textsf{Adapter-TST}'s ability to perform TST well and efficiently with fewer training parameters.

Table \ref{tbl:multi_output_results} shows the performance of the proposed \textsf{Adapter-TST} model and Style Transformer baselines on the multiple stylistic attribute output task. The \textsf{Adapter-TST} model achieves the best G-score over all four datasets while modeling multiple attributes with different adapters at the same time. We observe that \textsf{Adapter-TST} transfer tense attributes well and outperforms the baselines on three datasets. However, modeling multiple attributes simultaneously is a more challenging task. The proposed \textsf{Adapter-TST} model has an acceptable performance gap compared to the Style Transformer model on the other attribute transfer accuracy. \textsf{Adapter-TST} can generate fluent and complete sentences while preserving the original content; thus it can outperform the baselines on content preservation and fluency. Furthermore, it is important to note that multiple Style Transformers are trained for the multiple-attribute TST tasks, making it computationally inefficient and expensive compared to \textsf{Adapter-TST}. 

\begin{table*}[t]
\small
\centering
\begin{tabular}{cc|ccccc}
\hline
&& \multicolumn{5}{c}{Tense-Voice}\\
\textbf{Model} & \textbf{Attribute} & \textbf{Tense}  & \textbf{Voice}  & \textbf{BS} & \textbf{PPL} & \textbf{G}  \\
\hline
Style Transformer & Tense & 90.0 & - & 3.72 & 3.23 & 10.26 \\
Style Transformer & Voice & - & \textbf{74.0} & 2.16 & 2.42 & 7.29 \\
\textsf{Adapter-TST} (ours) & Tense+Voice & \textbf{99} & 67.0 & \textbf{3.74} & \textbf{3.58} & \textbf{10.35} \\
\hline
Sequential Style Transformer & Tense+Voice & 81.0 & \textbf{85.0} & 2.56 & 2.88 & 8.49\\
\textsf{Adapter-TST} (ours) & Tense+Voice & \textbf{93.0} & 82.0 & \textbf{3.19} & \textbf{3.00} & \textbf{9.43}\\
\hline
\end{tabular}
\caption{Human evaluation results of models on both multiple stylistic attribute outputs and compositional editing tasks. The best performances are \textbf{bold}.}
\label{tbl:human_results}
\end{table*}

\begin{table*}[t]
\small
\centering
\begin{tabular}{lp{4cm}p{4cm}p{4cm}}
\hline
\textbf{Target Style} & \textbf{Source Sentence} & \textbf{Style Transformer} & \textbf{\textsf{Adapter-TST}} \\
\hline
Future & The plan lacked a withdrawal timetable.  & The plan \red{will} lack be had by the buy-out group. & The plan \blue{will have} a withdrawal timetable. \\
Past & Some issues will be helped by higher earnings. & Some issues \red{were} helped by higher earnings by some issues. & Some issues \blue{were} helped by higher earnings. \\
Present & And he will question the white house dedication. & And he \red{question} the white house and he & And he \blue{says} the white house dedication. \\
Future+passive & Litigation sciences doesn't make moral distinctions. & Litigation transportation \red{will not} make fuel had and this teaches us and no she \red{will be had by} either. & Moral distinctions \blue{will be done by} Litigation sciences. \\
Past+Active & Third high yields are offered by them. & Third nutmeg yields \red{listed} night board them third period earnings stage  & Third high yields \blue{offered to} them. \\
\hline
\hline
\end{tabular}
\caption{Qualitative results for transfer to different target style combination across different models. Different colors highlight the transferred segments contributing to the target style.}
\label{tbl:case_study}
\end{table*}

To demonstrate that the attribute-specific adapters capture the corresponding attribute information, we evaluate the proposed \textsf{Adapter-TST} model on the compositional editing task. Note that the parameters of adapters trained in the multiple stylistic attribute outputs task are reloaded, and the connection method is changed to \textit{Stack} for compositional editing. Table \ref{tbl:compsotional_results} shows the performance of the \textsf{Adapter-TST} and Sequential Style Transformer on the compositional editing task. The \textsf{Adapter-TST} model achieves the highest G-score across four datasets, similar to the results obtained in the multiple stylistic attribute output task. We observe that the average G-score of the multiple stylistic attribute outputs task is 2.24, significantly higher than compositional editing's average G-score of 1.89. The difference in the average G-score highlights the challenge of the compositional editing task. Interestingly, \textsf{Adapter-TST} achieves comparable performance on style transfer accuracy over attributes, indicating that the attribute-specific adapters effectively capture the stylistic attributes. 

\subsection{Human Evaluation}

A human-based evaluation study was conducted to assess the performance of the \textsf{Adapter-TST} model in handling multiple-attribute TST tasks. Specifically, 200 sentences were randomly sampled from the Tense-Voice dataset. Both multiple stylistic attribute outputs and compositional editing are performed using \textsf{Adapter-TST} and the baselines on the sampled sentences. The generated sentences were then evaluated by two linguistic researchers using the three criteria employed for automated evaluation. We evaluate the \textit{transfer strength} by asking the evaluators to indicate whether the generated sentences are in the target attribute value (i.e., future tense, passive voice) using a binary true/false indicator. For \textit{content presentation}, the evaluators are instructed to rate the amount of content preserved in the generated sentences using a 5-point Likert scale, where 1 represents no content preserved, and 5 represents all content preserved. For fluency, the evaluators are instructed to rate the fluency of the transferred sentences using a 5-point Likert scale, where 1 represents unreadable with too many grammatical errors, and 5 represents a perfect and fluent sentence. To minimize biases, the model names are not displayed, and the order of the models is shuffled when displaying the generated sentences. This ensures that the evaluators are unaware of which model generated the sentences that they are evaluating.



\subsection{Human Evaluation Results}
Table \ref{tbl:human_results} shows the evaluation results. The style transfer accuracy of the models was computed using the binary feedback from the evaluators. The average scores for the criteria of content preservation and fluency were calculated using the 5-point Likert scores. \textsf{Adapter-TST} is observed to outperform the baselines in content preservation, fluency, and G-score. \textsf{Adapter-TST} is also rated to generate more syntactically sound and fluent sentences compared to the baselines. We can also observe that there is still a style transfer accuracy drop of \textsf{Adapter-TST} on attribute Voice when modeling multiple attributes at the same time. These results align with the automatic evaluations and demonstrate \textsf{Adapter-TST}'s effectiveness in performing multiple-attribute TST well and efficiently.

\section{Case Study}

We conduct case studies by presenting randomly sampled examples and the corresponding style transferred outputs of the \textsf{Adapter-TST} and the Style Transformer model. Table \ref{tbl:case_study} shows the example outputs on the Tense-Voice dataset. We observe that the proposed \textsf{Adapter-TST} has successfully transferred the style while preserving the content and sentence structure on multiple-attribute TST tasks. While Style Transformer has generated output sentences with grammatical errors, making it harder to judge if the style has been successfully transferred. Additionally, the Style Transformer performs poorly on the compositional editing task due to its inherent complexity. Albeit the difficulty of compositional editing, \textsf{Adapter-TST} is able to generate a fluent sentence that preserves the original content.

\section{Conclusion}
In this paper, we introduced a parameter-efficient framework, \textsf{Adapter-TST} with different neural adapters to capture different attribute information for multiple-attribute TST tasks. During training, the original parameters of BART-large were frozen, and only the adapters' parameters were optimized to relax the dependence on computational resources and supervised data. We conducted extensive experiments on traditional sentiment transfer and multiple-attribute TST tasks. The automatic and human-based evaluation results showed that the attribute-specific adapters in \textsf{Adapter-TST} is able to capture relevant stylistic attributes to transfer the style while preserving the original content successfully. Our case studies also demonstrated that \textsf{Adapter-TST} was able to generate high-quality text in the target style. For future work, we will continue to improve TST models' ability to model multiple attributes in terms of quality and efficiency. We will also explore plugging \textsf{Adapter-TST} on other PLMs and evaluate its effectiveness.

\section{Limitations}
This work has two limitations. First, there is a style transfer accuracy reduction on one of the attributes, while the proposed model models multiple attributes simultaneously. Explorations on improving TST models' ability to handle multiple-attribute TST tasks and the dependency among attributes are potential directions in this field. Second, even though we have frozen the parameters of the pre-trained BART-large model to improve parameter efficiency, we still need to run BART-large model to extract representations for performing TST tasks. 

\section{Ethics Statement}
The ethical implications of using large language models trained on data containing unchecked biases are acknowledged. As with any generative task, style transfer also has the potential for misuse, such as fact distortion, plagiarism, etc. The paper aims to demonstrate the academic utility of the proposed framework. This solution must be paired with strict checks for misrepresentation, offensiveness, and bias to adhere to ethical standards.

\bibliographystyle{acl_natbib}
\bibliography{refers}




\end{document}